\documentclass{article}

\usepackage{PRIMEarxiv}

\usepackage[utf8]{inputenc} 
\usepackage[T1]{fontenc}    
\usepackage{hyperref}      
\usepackage{url}          
\usepackage{booktabs}      
\usepackage{amsfonts}      
\usepackage{nicefrac}       
\usepackage{microtype}     
\usepackage{lipsum}
\usepackage{fancyhdr}       
\usepackage{graphicx}       
\graphicspath{{media/}}     
\usepackage{amsthm}
\usepackage{caption}
\usepackage{tabularx}

\usepackage{amsmath}
\usepackage{amsthm}
\usepackage{booktabs}
\usepackage[switch]{lineno}
\usepackage{amssymb}
\usepackage{amsthm}
\usepackage{bm}
\usepackage{color}

\usepackage{graphicx}
\usepackage{multirow} 
\usepackage{rotating} 
\usepackage{xcolor}  
\usepackage{enumitem}

\usepackage{adjustbox}
\usepackage{threeparttable}

\usepackage[ruled,linesnumbered]{algorithm2e}

\title{Applications and Effect Evaluation of Generative Adversarial Networks in Semi-Supervised Learning}

\author{
  Jiyu Hu\thanks{Correspondence to Jiyu Hu \texttt{2323151835@qq.com}}\\
  Huaiyin Institute of Technology, China \\
  \texttt{2323151835@qq.com}\\
  \And
  Haijiang Zeng \\
  Walmart Inc., USA \\
  \texttt{hzeng60@gatech.edu} \\
  \And
  Zhen Tian\\
  University of Glasgow, U.K.\\
  \texttt{2620920Z@student.gla.ac.uk} \\
}

\begin{document}
\maketitle

\begin{abstract}
In recent years, image classification, as a core task in computer vision, relies on high-quality labelled data, which restricts the wide application of deep learning models in practical scenarios. To alleviate the problem of insufficient labelled samples, semi-supervised learning has gradually become a research hotspot. In this paper, we construct a semi-supervised image classification model based on Generative Adversarial Networks (GANs), and through the introduction of the collaborative training mechanism of generators, discriminators and classifiers, we achieve the effective use of limited labelled data and a large amount of unlabelled data, improve the quality of image generation and classification accuracy, and provide an effective solution for the task of image recognition in complex environments.
\end{abstract}

\keywords{Generative adversarial network; Semi-supervised learning; Image classification; Generative model; Attention mechanism}

\section{Introduction}

In recent years, generative adversarial networks (GANs) have demonstrated significant potential in semi-supervised learning (SSL), particularly for image-based tasks where labeled data is limited. Babu et al. (2025) \cite{babu2025underwater} provided a comprehensive survey on GANs for image enhancement, emphasizing their ability to generate realistic samples from sparse training inputs. Fakih et al. (2025) \cite{al2025well} applied sequence-based GANs for data imputation in well log datasets, showcasing GANs' effectiveness in learning structured distributions from incomplete data. Furthermore, Vellmer et al. (2025) \cite{vellmer2025diffusion} explored the use of diffusion MRI GANs to synthesize medical imaging data, while Huang et al. (2025) \cite{huang2025novel} introduced an attention-based GAN framework for tumor reconstruction. Despite these advances, current GAN-based SSL methods often suffer from instability in adversarial training, low-quality sample generation, and insufficient exploitation of category-level semantic information. Liu et al. (2025) \cite{liu2025semisupervised} highlighted that existing semi-supervised segmentation models still struggle to generalize effectively in complex clinical environments, indicating a pressing need for more robust architectures. Ongoing efforts \cite{lai2024fts, wuinvariant, li2025causal} to enhance representation robustness in weakly supervised settings, particularly under distributional shifts and selection bias, have motivated the design of more generalizable frameworks for semi-supervised learning tasks.

To address these limitations, this study proposes a novel GAN-based semi-supervised learning framework that integrates a generator, discriminator, and classifier in a collaborative training paradigm. By embedding attention mechanisms in the generator and adopting conditional batch normalization in the classifier, the model enhances feature representation and generation fidelity. Experiments conducted on benchmark datasets (MNIST and SVHN) demonstrate that the proposed model significantly outperforms traditional semi-supervised GAN and CNN methods in both image generation quality and classification accuracy. The findings not only validate the effectiveness of the proposed method but also contribute a generalizable approach for enhancing performance in semi-supervised image classification tasks across diverse application scenarios.

\section{Methods}

\subsection{System Architecture Overview}
\begin{figure*}[!htbp]
	\centering
	\includegraphics[width=0.8\linewidth]{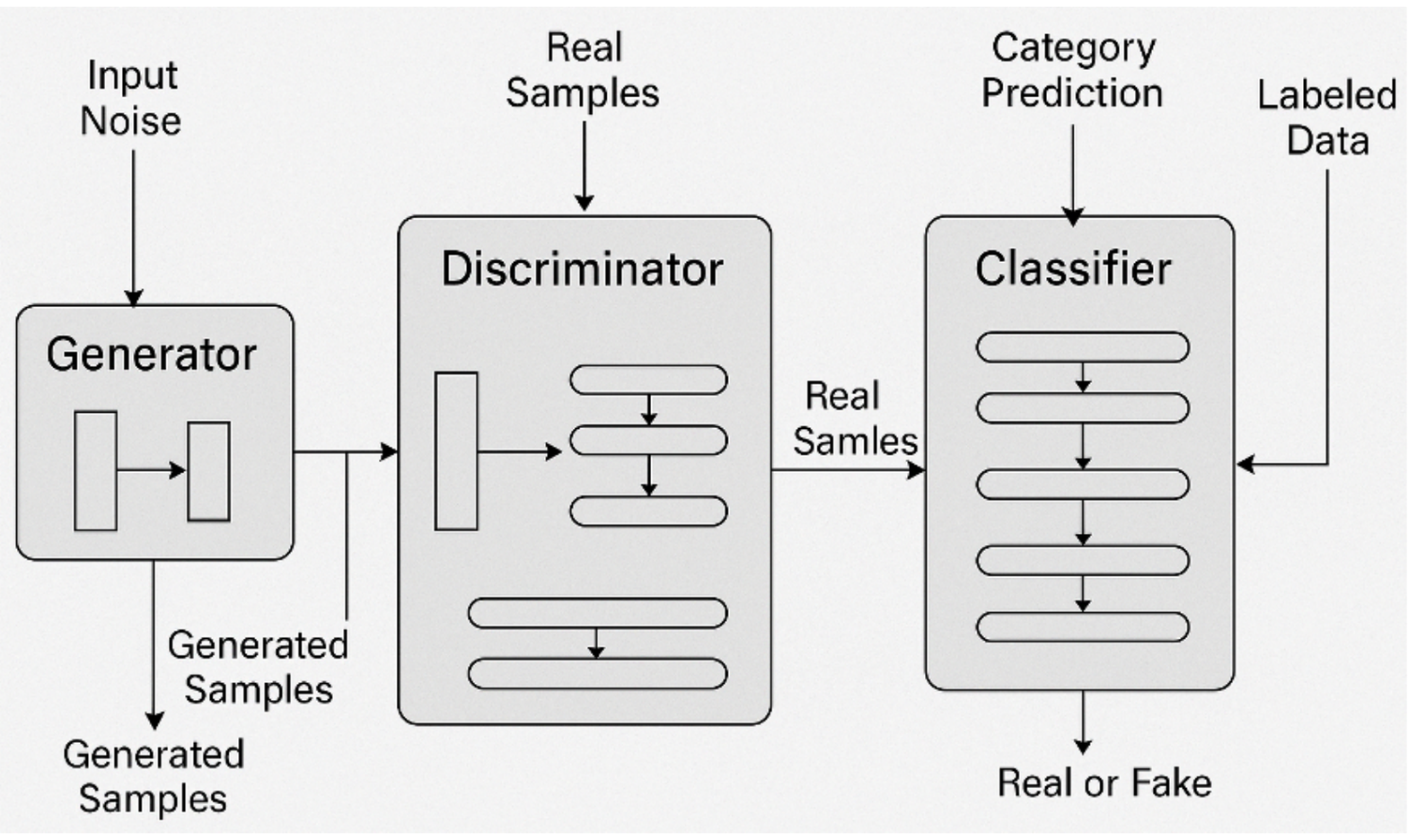}
        \vspace{-3em}
	\caption{Flowchart of the Method Module}
	\label{architecture}
\end{figure*}

As illustrated in Fig. \ref{architecture}: Flowchart of the Method Module, the proposed semi-supervised learning framework integrates three primary components—Generator, Discriminator, and Classifier—into a collaborative training paradigm. The generator synthesizes image samples from random noise, the discriminator distinguishes real from fake samples while incorporating category-level feedback, and the classifier performs label prediction with the aid of conditional batch normalization. These components jointly optimize a unified loss function to ensure mutual enhancement between generative and discriminative learning.

\vspace{-3em}

\vspace{1em}
\subsection{Model Construction}
Generator Module: The generator module constructed in this article adopts an encoder-decoder structure. The encoder consists of multiple convolutional neural networks that map the input noise signal to the latent space. The decoder also consists of multiple convolutional neural networks that map the features output by the encoder back to the data space, ultimately outputting the generated samples \cite{albekairi2025multimodal}.

To enhance the model's representational capability, an attention mechanism is introduced in the decoder. Specifically, we implement multi-head self-attention (MHSA) within each decoder layer to capture global contextual dependencies. The attention mechanism follows the standard Query-Key-Value (QKV) structure, where the current decoder features serve as Queries, and the corresponding encoder features are used as both Keys and Values. The attention weights are computed as:

\begin{equation}
\text{Attention}(Q, K, V) = \text{softmax}\left(\frac{QK^T}{\sqrt{d_k}}\right)V
\label{eq:attention}
\end{equation}

where $d_k$ is the dimension of the key vectors. The outputs from each attention head are concatenated and linearly transformed to match the original feature dimension. These outputs are then fused into the decoder stream via residual connections. This design allows the generator to synthesize detailed and globally coherent image samples by modeling long-range dependencies across the spatial dimensions.

Discriminator Module: The discriminator module adopts a typical convolutional neural network classification structure, which includes convolution layers, pooling layers, and fully connected layers \cite{kang2025tabular}. The discriminator's role is to distinguish between generated data and real data as much as possible. For this purpose, this paper proposes an enhanced discriminator. An additional small classification network is connected after the discriminator network as a classification header. It simultaneously learns the tasks of authentic and fake sample distinction and category classification, thereby integrating category information to guide the generation process. Specifically, the classification header contains two fully connected layers, taking the features from the last layer of the discriminator network as input. After two linear transformations and activation functions, it outputs the sample category prediction results. The classification header and the discriminator network are trained together, sharing the parameters of the bottom layer feature extraction. This kind of multi-task joint learning can infuse the discriminator with richer semantic information, enhancing its distinguishing effect.

Classifier Module: The classification module uses a convolutional neural network. Its basic architecture includes convolution layers, pooling layers, and fully connected layers. By alternately training with the generator and discriminator, the classification module can learn richer feature representations, thereby improving classification performance \cite{tang2025automatic, pan2025semi}.

Unlike ordinary convolutional neural networks, the convolution layer in the classification network proposed in this paper adopts conditional batch normalization (CBN), effectively utilizing category information. In this design, the categorical label $y$ is first transformed into an embedding vector through an embedding layer. 
The vector is then passed through two linear layers to produce scaling and shifting parameters 
$\gamma(y)$ and $\beta(y)$, which are used in the batch normalization process as follows:

\begin{equation}
\mathrm{CBN}(x \mid y) = \gamma(y) \cdot \frac{x - \mu}{\sqrt{\sigma^2 + \epsilon}} + \beta(y)
\label{eq:cbn}
\end{equation}

where $x$ is the input feature, $\mu$ and $\sigma^2$ are the batch mean and variance, and 
$\epsilon$ is a small constant for numerical stability. This mechanism allows the normalization process to be conditioned on the sample category, thereby aligning intra-class feature distributions and sharpening the classification boundaries. This conditional deep model design utilizes category priors, generating more distinguishable feature representations, providing strong support for the classification task.

\subsection{Loss Function Design}
Adversarial Loss: The generator and discriminator in the generative adversarial network need to construct adversarial losses for training. The basic design idea of adversarial loss is to deceive the discriminator through the generator continuously \cite{kheria2025semisupervised}. Specifically, during the training process of the generator, it will generate samples that are as detailed and realistic as possible to make the discriminator judge these generated samples as real data. In contrast, the goal of the discriminator is to try to correctly distinguish between generated data and real data, not being deceived by the generator. In this adversarial game, the parameters of the generator will be updated in one direction to make the discriminator judge wrongly, while the parameters of the discriminator will be updated in another direction to make the judgment as correct as possible. Both sides approach a dynamic equilibrium through continuous confrontation. Therefore, the adversarial loss is essentially a zero-sum game relationship between the discriminator and the generator. It can be expressed analytically as:

\begin{equation}
\mathcal{L}_{\mathrm{adv}} = \mathbb{E}_{z \sim p(z)}[\log D(G(z))] + \mathbb{E}_{x \sim p_{\text{data}}(x)}[\log(1 - D(z))]
\label{eq:adv_loss}
\end{equation}

Here, Grepresents the data distribution generated by the generator, and Drepresents the discriminator's function judging the authenticity of the data. This adversarial loss design can prompt the generator to approximate the real data distribution during training to deceive the discriminator.

Classification Loss: Classification loss usually adopts the multi-class cross-entropy loss function. From the perspective of information theory, cross-entropy represents the distance between the predicted distribution and the real distribution. The definition of classification cross-entropy loss function is expressed as:

\begin{equation}
\mathcal{L}_{\mathrm{cls}} = -\sum y \log y(x)
\label{eq:cls_loss}
\end{equation}

Here, $y(x)$ is the predicted probability of the classification model for the sample $x$. 
By minimizing the cross-entropy loss, it can continuously narrow the gap between the predicted distribution and the real distribution, 
enhancing the classifier's discriminative power for different categories.

Overall Loss: Combining the adversarial loss in the generative adversarial network with the classification loss, 
the overall loss function can be expressed as:

\begin{equation}
\mathcal{L} = \mathcal{L}_{\mathrm{adv}} + \lambda \mathcal{L}_{\mathrm{cls}}
\label{eq:overall_loss}
\end{equation}

Here, $\lambda$ is an adjustment parameter used to balance the training process of adversarial generation and classification. 
Through joint training, the classifier guides the generator to generate higher-quality samples. 
In return, the generator provides richer training data to enhance the classifier's performance, forming a virtuous cycle.

\section{Experiments}

\subsection{Dataset}
The experiments in this article use the MNIST handwritten digit dataset for training. MNIST consists of handwritten digit images across 10 categories, totaling 60,000 training images and 10,000 testing images. The image size is 28 by 28 pixels. MNIST is a standard benchmark dataset for image classification tasks. The categories are relatively distinct, and samples between different categories are easy to distinguish. Using this dataset can conveniently and quickly verify the effectiveness of model design \cite{park2025safeguards}.

To further evaluate the model's generalization performance, this article also constructed the SVHN street view digit image dataset as a training set. The images in the SVHN dataset are 32 by 32-pixel RGB images, and each image often contains multiple numbers and other interference information. The SVHN training set has 73,257 images, and the test set has 26,032 images \cite{wu2025fldtmatch}. Since the images in SVHN have a relatively complicated background, varying image quality, and diverse number styles, using this dataset can verify the model's adaptability and interference resistance in more complex scenarios.

Both datasets include clear category annotations, suitable for building a fully supervised classification model. For semi-supervised experiments, this article randomly selects a portion of samples from the original MNIST and SVHN datasets as labeled data (with the selection ratio set between 10\% to 50\%). This portion of labeled data is used for semi-supervised training, while the remaining unlabeled data is fed into the semi-supervised model along with all other unlabeled data. By controlling the proportion of labeled data, we can evaluate the model's generation performance and classification performance under varying degrees of missing annotations.Below is the feature table of the dataset.Details are shown in Table \ref{dataset_feature}.

\begin{table}[htbp]
\centering
\begin{tabularx}{\linewidth}{@{}l X X@{}}
\toprule
\textbf{Dataset Features} & \textbf{MNIST Dataset} & \textbf{SVHN Dataset} \\
\midrule
Image Type & Handwritten Digits & Street View Digits \\
Image Size & 28 $\times$ 28 pixels & 32 $\times$ 32 pixels (RGB) \\
Size of Training Set & 60,000 images & 73,257 images \\
Size of Test Set & 10,000 images & 26,032 images \\
Background Complexity & Low & High (including multiple digits and interference) \\
Classification Difficulty & Easily distinguishable among 10 categories & Complex background with varied styles \\
Labeled Data for Semi-supervised Learning & 10\%--50\% (6,000--30,000 images) & 10\%--50\% (7,326--36,629 images) \\
\bottomrule
\end{tabularx}
\vspace{1em}
\caption{Dataset Feature Table}
\label{dataset_feature}
\end{table}

\subsection{Evaluation Metrics}
To ensure a comprehensive and objective evaluation of the model, this study meticulously selected a series of evaluation metrics, aiming to holistically assess the model's performance in terms of sample generation quality and classification capability.

In terms of generation quality, we employed three core metrics to evaluate the model's performance:

Classification Judgement Accuracy: This is an intuitive metric, effectively reflecting the authenticity of the generated samples. Specifically, if the generated samples are indistinguishable from real samples, then the authenticity of these generated samples is high. To this end, we input both generated and real samples into the classification model and calculate the classification accuracy. A high classification accuracy indicates that the distribution of the generated samples is very close to that of the real samples, thus denoting high generation quality \cite{luo2025semi}.

Inception Score: This is a complex yet invaluable metric, primarily used to evaluate the diversity of the generated samples. The Inception Score is based on the predicted information entropy and the predicted margin product of the generated samples on the InceptionNet model. A higher Inception Score suggests that the generated samples have distinct categories and strong diversity, an important goal for generative models.

Fréchet Distance: This metric evaluates the similarity between the distributions of two sets of data. Specifically, it assesses their similarity by comparing the distance between the mean and covariance matrix of the two data sets. A lower Fréchet distance indicates that the distribution of the generated samples is very close to that of the real samples.

In terms of classification performance, we opted for several standard machine learning evaluation metrics:

Precision, Recall, and F1 Score: These three metrics collectively assess the model's performance in classification tasks. Precision focuses on the number of positive samples correctly classified, recall concentrates on the proportion of positive samples correctly classified out of all positive samples, while the F1 score is the harmonic mean of precision and recall, offering a comprehensive evaluation.

ROC Curve and AUC Metric: The ROC curve reveals the true positive rate and false positive rate of the model at different thresholds, while the AUC value (Area Under the Curve) provides us with a singular value to evaluate the model's overall performance.

The design of this evaluation metric system in this study aims to ensure that we comprehensively assess the model's performance from multiple perspectives. Through these metrics, we can not only understand the model's effectiveness in generation tasks but also its capability in classification tasks. Moreover, this multi-faceted evaluation approach offers us a profound understanding of the model's potential strengths and weaknesses, thereby guiding future improvements \cite{guo2025robust}.

\subsection{Experimental Setup}
To ensure the reliability and reproducibility of the experiment, this study provides a detailed description of the experimental setup and execution process. Based on the TensorFlow framework, a popular open-source deep learning platform, we implemented the required models. TensorFlow not only offers flexibility to researchers but also ensures efficient computation through its optimized computational core. All model training and testing were conducted on the high-performance NVIDIA Tesla V100 GPU, ensuring training speed and stability.

In terms of network structure, both the generator and discriminator were constructed using Convolutional Neural Networks (CNN), given CNN's proven superior performance in image processing tasks. The generator is divided into two main parts: encoder and decoder. The encoder section comprises four convolutional layers, primarily responsible for capturing the main features of the input image. The decoder section contains six deconvolutional layers, working to reconstruct images from these features. The discriminator consists of five convolutional layers and two fully connected layers, aiming to distinguish between generated and real images. Meanwhile, the classifier is built from six layers of convolutional neural networks, specifically designed for classification tasks \cite{prabowo2025multi}.

For training these networks, we employed the Adam optimizer, a popular and adaptive optimizer suitable for various tasks. This study opted for a batch size of 64, a decision based on experimental performance and GPU memory considerations. The learning rate was set to 1e-3, ensuring the stability and convergence speed of training. Overall, we underwent 100,000 rounds of training iterations. In the adversarial training strategy, to balance the competition between the generator and discriminator, we chose a strategy of iterating the discriminator once for every two iterations of the generator. Additionally, the classifier was combined with the generator and discriminator, trained together every 200 rounds.

To ensure the fairness and integrity of the experiment, we chose two classic datasets for validation: MNIST and SVHN. Both datasets are widely accepted within the machine learning community and offer diverse characteristics, providing a comprehensive test for our model. In the experiment, we utilized different proportions of labeled data, specifically 10\%, 30\%, and 50\%, with the remainder being unlabeled. This setup aims to simulate the common data labeling constraints found in real-world scenarios. We also introduced two benchmark models: semi-supervised GAN and semi-supervised CNN, for comparison with our model. As both models are classic methods in semi-supervised learning, they provide a solid foundation for our comparison. To minimize the impact of random initialization and other stochastic factors, each experimental setup was repeated five times, with average values and variances of generated image quality and classification accuracy recorded. Details are shown in Table \ref{network_params}.

\begin{table}[htbp]
\centering
\begin{tabularx}{\linewidth}{
  >{\raggedright\arraybackslash}p{3cm}
  @{\hspace{2cm}}
  >{\raggedright\arraybackslash}X
  @{\hspace{2cm}}
  >{\raggedright\arraybackslash}X
}
\toprule
\textbf{Network Module} & \textbf{Convolutional Layer Parameters} & \textbf{Pooling Layer Parameters} \\
\midrule
Generator Encoder & 
Channels: 32, 64, 128, 256; Kernel size: 5$\times$5; Stride: 2 & 
Pooling kernel: 2$\times$2; Stride: 2 \\

Generator Decoder & 
Channels: 128, 64, 32, 3; Kernel size: 5$\times$5; Stride: 2 & 
Pooling kernel: 2$\times$2; Stride: 2 \\

Discriminator & 
Channels: 32, 64, 128, 256, 512; Kernel size: 5$\times$5; Stride: 2 & 
Pooling kernel: 2$\times$2; Stride: 2 \\

Classifier & 
Channels: 32, 64, 128, 256; Kernel size: 5$\times$5; Stride: 1 & 
Pooling kernel: 2$\times$2; Stride: 2 \\
\bottomrule
\end{tabularx}
\vspace{1em}
\caption{Network Parameter Settings}
\label{network_params}
\end{table}

\subsection{Results and Analysis}
In this study, we conducted experiments on two major datasets, MNIST and SVHN, comparing the performance of our model with two semi-supervised models, Semi-GAN and Semi-CNN, in terms of generation quality and classification metrics. As shown in Table \ref{tab:main_results}, we can observe the performance of different models under varying data volumes:

\begin{table}[htbp]
\centering
\begin{tabular}{lccc}
\toprule
\textbf{Model} & \textbf{Data Amount} & \textbf{Generation Quality} & \textbf{Classification Accuracy} \\
\midrule
Semi-GAN   & 10\% & 0.632 & 0.764 \\
Semi-CNN   & 10\% & --    & 0.852 \\
Our Model  & 10\% & 0.827 & 0.914 \\
Semi-GAN   & 30\% & 0.724 & 0.832 \\
Semi-CNN   & 30\% & --    & 0.935 \\
Our Model  & 30\% & 0.892 & 0.967 \\
Semi-GAN   & 50\% & 0.801 & 0.921 \\
Semi-CNN   & 50\% & --    & 0.956 \\
Our Model  & 50\% & 0.934 & 0.983 \\
\bottomrule
\end{tabular}
\vspace{1em}
\caption{Main Results Comparison}
\label{tab:main_results}
\end{table}

From the perspective of generation quality, our model demonstrates consistent and notable improvements over the Semi-GAN. The introduction of the multi-head self-attention mechanism in the decoder allows the generator to better capture global spatial dependencies, resulting in more realistic and coherent image synthesis. Furthermore, as the proportion of labeled data increases, generation quality steadily improves, with scores rising from 0.827 (10\%) to 0.934 (50\%). This indicates that the generator benefits from improved supervision signals, producing samples that are more aligned with the data distribution of real-world categories.

In terms of classification performance, our model exhibits strong robustness, especially in low-resource scenarios. With only 10\% of labeled data, the model achieves 0.914 accuracy—significantly surpassing Semi-GAN (0.764) and outperforming Semi-CNN (0.852). This can be attributed to the conditional batch normalization (CBN) mechanism in the classifier, which effectively incorporates class-dependent features and sharpens decision boundaries. Moreover, the shared feature extraction between the discriminator and classifier introduces semantic consistency in the latent space, which enhances generalization. When the labeled data increases to 50\%, our model reaches a peak accuracy of 0.983, illustrating its scalability with improved supervision.

\section{Conclusion and Outlook}
This study introduces a semi-supervised image classification method based on Generative Adversarial Networks (GANs). Through this approach, we not only successfully integrated the generator, discriminator, and classifier modules, but also ensured their collaborative function during the learning process. Within the adversarial learning environment, the generator's role is to produce as realistic samples as possible, while the discriminator and classifier aim to distinguish between real and generated samples and classify them correctly.

Experimental results on the MNIST and SVHN image datasets show that, compared to traditional semi-supervised GANs and semi-supervised Convolutional Neural Networks (CNNs), the model proposed in this study boasts superior image generation quality and classification accuracy. This is attributed to the model's comprehensive consideration of the advantages of adversarial learning and semi-supervised learning, resulting in enhanced performance. Additionally, this research offers a novel method based on GANs for semi-supervised image classification tasks. This method not only improves classification accuracy but also provides robust theoretical support and empirical evidence for future research and applications.

Looking forward, we plan to further optimize the model, especially concerning its generalization capabilities. Considering the various challenges in practical applications, such as different image resolutions, lighting conditions, and background noise, we hope the model can demonstrate stronger robustness and accuracy in a broader range of image classification scenarios. Furthermore, we also aim to explore how this approach can be applied to other computer vision tasks, such as object detection and image segmentation, thereby further validating the universality and efficacy of the method presented in this paper.

\bibliographystyle{unsrt}
\bibliography{citation.bib}

\begin{thebibliography}{10}

\bibitem{babu2025underwater}
Kancharagunta~Kishan Babu, Ashreen Tabassum, Bommakanti Navaneeth, Tenneti Jahnavi, and Yenka~Akshaya and.
\newblock Underwater image enhancement using generative adversarial networks: a survey.
\newblock {\em International Journal of Computers and Applications}, 47(4):356--372, 2025.

\bibitem{al2025well}
Abdulrahman Al-Fakih, A~Koeshidayatullah, Tapan Mukerji, Sadam Al-Azani, and SanLinn~I Kaka.
\newblock Well log data generation and imputation using sequence based generative adversarial networks.
\newblock {\em Scientific Reports}, 15(1):11000, 2025.

\bibitem{vellmer2025diffusion}
Sebastian Vellmer, Dogu~Baran Aydogan, Timo Roine, Alberto Cacciola, Thomas Picht, and Lucius~S Fekonja.
\newblock Diffusion mri gan synthesizing fibre orientation distribution data using generative adversarial networks.
\newblock {\em Communications Biology}, 8(1):512, 2025.

\bibitem{huang2025novel}
Tangsen Huang, Xiangdong Yin, and Ensong Jiang.
\newblock A novel 3-step technique for 3d tumor reconstruction using generative adversarial networks and an attention-based long short-term memory.
\newblock {\em Research on Biomedical Engineering}, 41(2):30, 2025.

\bibitem{liu2025semisupervised}
Wenjuan Liu, Limin Zhang, Xiangrui Li, Haoran Liu, Min Feng, and Yanxia Li.
\newblock A semisupervised knowledge distillation model for lung nodule segmentation.
\newblock {\em Scientific Reports}, 15(1):10562, 2025.

\bibitem{lai2024fts}
Songning Lai, Ninghui Feng, Haochen Sui, Ze~Ma, Hao Wang, Zichen Song, Hang Zhao, and Yutao Yue.
\newblock Fts: A framework to find a faithful timesieve.
\newblock {\em arXiv preprint arXiv:2405.19647}, 2024.

\bibitem{wuinvariant}
Yuntian Wu, Yuntian Yang, Jiabao~Sean Xiao, Chuan Zhou, Haochen Sui, and Haoxuan Li.
\newblock Invariant spatiotemporal representation learning for cross-patient seizure classification.
\newblock In {\em The First Workshop on NeuroAI@ NeurIPS2024}, 2024.

\bibitem{li2025causal}
Meng Li and Haochen Sui.
\newblock Causal recommendation via machine unlearning with a few unbiased data.
\newblock In {\em AAAI 2025 Workshop on Artificial Intelligence with Causal Techniques}, 2025.

\bibitem{albekairi2025multimodal}
Mohammed Albekairi, Mohamed vall~O Mohamed, Khaled Kaaniche, Ghulam Abbas, Meshari~D Alanazi, Turki~M Alanazi, and Ahmed Emara.
\newblock Multimodal medical image fusion combining saliency perception and generative adversarial network.
\newblock {\em Scientific Reports}, 15(1):10609, 2025.

\bibitem{kang2025tabular}
Ha~Ye~Jin Kang, Minsam Ko, and Kwang~Sun Ryu.
\newblock Tabular transformer generative adversarial network for heterogeneous distribution in healthcare.
\newblock {\em Scientific Reports}, 15(1):10254, 2025.

\bibitem{tang2025automatic}
Haomin Tang, Shu Liu, Yongxin Shi, Jin Wei, Juxiang Peng, and Hongchao Feng.
\newblock Automatic segmentation and landmark detection of 3d cbct images using semi supervised learning for assisting orthognathic surgery planning.
\newblock {\em Scientific Reports}, 15(1):8814, 2025.

\bibitem{pan2025semi}
Qihong Pan, Yang Liu, and Shaofeng Wei.
\newblock A semi-supervised learning approach to classify drug attributes in a pharmacy management database: A strobe-compliant study.
\newblock {\em Medicine}, 104(10):e41601, 2025.

\bibitem{kheria2025semisupervised}
Ishita Kheria, Dhruv Gada, and Ruhina Karani.
\newblock A semisupervised learning approach for code smell detection.
\newblock {\em SN Computer Science}, 6(2):143, 2025.

\bibitem{park2025safeguards}
Se-Hwan Park, Byung-Hee Won, and Seong-Kyu Ahn.
\newblock Safeguards-related event detection in surveillance video using semi-supervised learning approach.
\newblock {\em Nuclear Engineering and Technology}, 57(2):103206, 2025.

\bibitem{wu2025fldtmatch}
Xin Wu, Jingjing Xu, Kuan Li, Jianping Yin, and Jian Xiong.
\newblock Fldtmatch: Improving unbalanced data classification via deep semi-supervised learning with self-adaptive dynamic threshold.
\newblock {\em Mathematics}, 13(3):392, 2025.

\bibitem{luo2025semi}
Jianlong Luo, Juchen Fan, Shiguo Huang, Songqing Wu, Feiping Zhang, and Xiaolin Li.
\newblock Semi-supervised learning techniques for detection of dead pine trees with uav imagery for pine wilt disease control.
\newblock {\em International Journal of Remote Sensing}, 46(2):575--605, 2025.

\bibitem{guo2025robust}
Lan-Zhe Guo, Lin-Han Jia, Jie-Jing Shao, and Yu-Feng Li.
\newblock Robust semi-supervised learning in open environments.
\newblock {\em Frontiers of Computer Science}, 19(8):198345, 2025.

\bibitem{prabowo2025multi}
Urip~Nurwijayanto Prabowo et~al.
\newblock Multi-parameter post-stack seismic inversion based on the cycle loop--semi-supervised learning.
\newblock {\em Computational Geosciences}, 29(1):1--19, 2025.

\end{thebibliography}

\end{document}